\documentclass{article}

\usepackage{amsmath,amsfonts,bm}

\def\eqref#1{equation~\ref{#1}}

\def\1{\bm{1}}

\DeclareMathAlphabet{\mathsfit}{\encodingdefault}{\sfdefault}{m}{sl}
\SetMathAlphabet{\mathsfit}{bold}{\encodingdefault}{\sfdefault}{bx}{n}

\newcommand{\R}{\mathbb{R}}

\usepackage{microtype}
\usepackage{graphicx}
\usepackage{subfigure}
\usepackage{booktabs} 

\usepackage{hyperref}

\usepackage[accepted]{icml2025}

\usepackage{amsmath}
\usepackage{amssymb}
\usepackage{mathtools}
\usepackage{amsthm}

\usepackage[capitalize,noabbrev]{cleveref}

\theoremstyle{plain}
\newtheorem{theorem}{Theorem}[section]
\newtheorem{proposition}[theorem]{Proposition}
\newtheorem{lemma}[theorem]{Lemma}
\newtheorem{corollary}[theorem]{Corollary}
\theoremstyle{definition}
\newtheorem{definition}[theorem]{Definition}
\newtheorem{assumption}[theorem]{Assumption}
\theoremstyle{remark}
\newtheorem{remark}[theorem]{Remark}

\usepackage{hyperref}
\usepackage{url}
\usepackage[utf8]{inputenc} 
\usepackage[T1]{fontenc}    
\usepackage{hyperref}       
\usepackage{url}            
\usepackage{booktabs}       
\usepackage{amsfonts}       
\usepackage{nicefrac}       
\usepackage{microtype}      
\usepackage{xcolor}         
\usepackage{makecell}
\usepackage{graphicx}
\usepackage{amsmath}
\usepackage{multirow}
\usepackage{wrapfig}
\usepackage{enumitem}
\usepackage{wrapfig}
\usepackage{float}
\usepackage{graphicx}
\usepackage{caption}
\usepackage{color}
\usepackage{colortbl}

\newcommand{\EQ}{\begin{eqnarray}}

\newcommand{\EN}{\end{eqnarray}}
\newcommand{\EQQ}{\begin{eqnarray*}}
\newcommand{\ENN}{\end{eqnarray*}}
\newcommand{\overlineray}{\begin{array} }

\newcommand{\Q}{{\mathbf Q}}
\newcommand{\K}{{\mathbf K}}
\newcommand{\V}{{\mathbf V}}
\newcommand{\X}{{\mathbf X}}
\newcommand{\barray}{\begin{array} }
\newcommand{\earray}{\end{array}}

\newcommand{\bremark}{\begin{remark} }
\newcommand{\eremark}{\end{remark}}
\newcommand{\btheorem}{\begin{theorem}}
\newcommand{\etheorem}{\end{theorem}}
\newcommand{\blemma}{\begin{lemma}}
\newcommand{\elemma}{\end{lemma}}
\newcommand{\bassumption}{\begin{assumption} }
\newcommand{\eassumption}{\end{assumption}}
\newcommand{\bcorollary}{\begin{corollary} }
\newcommand{\ecorollary}{\end{corollary}}
\newcommand{\bdefinition}{\begin{definition} }
\newcommand{\edefinition}{\end{definition}}
\newcommand{\bproposition}{\begin{proposition}}
\newcommand{\eproposition}{\end{proposition}}

\newcommand{\balgorithm}{\medskip\begin{algorithm} \rm}
\newcommand{\ealgorithm}{ \hfill \rule{1mm}{2mm}\medskip
\end{algorithm} }

\usepackage{microtype}
\usepackage{graphicx}
\usepackage{subfigure}
\usepackage{booktabs} 
\usepackage{paralist}

\usepackage[textsize=tiny]{todonotes}

\icmltitlerunning{LASP-2}

\begin{document}

\twocolumn[
\icmltitle{LASP-2: Rethinking Sequence Parallelism for Linear Attention and Its Hybrid}



\icmlsetsymbol{equal}{*}

\begin{icmlauthorlist}
\icmlauthor{Weigao Sun}{yyy}
\icmlauthor{Disen Lan}{yyy,xxx}
\icmlauthor{Yiran Zhong}{yyy}
\icmlauthor{Xiaoye Qu}{yyy}
\icmlauthor{Yu Cheng}{sch}
\end{icmlauthorlist}

\icmlaffiliation{yyy}{Shanghai AI Laboratory}
\icmlaffiliation{xxx}{South China University of Technology}
\icmlaffiliation{sch}{The Chinese University of Hong Kong}

\icmlcorrespondingauthor{Yu Cheng}{chengyu@cse.cuhk.edu.hk}


\icmlkeywords{Machine Learning, ICML}

\vskip 0.3in
]



\printAffiliationsAndNotice{}  

\begin{abstract}

Linear sequence modeling approaches, such as linear attention, provide advantages like linear-time training and constant-memory inference over sequence lengths. However, existing sequence parallelism (SP) methods are either not optimized for the right-product-first feature of linear attention or use a ring-style communication strategy, which results in lower computation parallelism, limits their scalability for longer sequences in distributed systems. In this paper, we introduce LASP-2, a new SP method to enhance both communication and computation parallelism when training linear attention transformer models with very-long input sequences. Compared to previous work LASP, LASP-2 rethinks the minimal communication requirement for SP on linear attention layers, reorganizes the whole communication-computation workflow of LASP. In this way, only one single AllGather collective communication is needed on intermediate memory states, whose sizes are independent of the sequence length, leading to significant improvements of both communication and computation parallelism, as well as their overlap. Additionally, we extend LASP-2 to LASP-2H by applying similar communication redesign to standard attention modules, offering an efficient SP solution for hybrid models that blend linear and standard attention layers. Our evaluation on a Linear-Llama3 model, a variant of Llama3 with linear attention replacing standard attention, demonstrates the effectiveness of LASP-2 and LASP-2H. Specifically, LASP-2 achieves training speed improvements of 15.2\% over LASP and 36.6\% over Ring Attention, with a sequence length of 2048K across 64 GPUs. The Code is released as a part of: \url{https://github.com/OpenSparseLLMs/Linear-MoE}.

\end{abstract}

\section{Introduction}

Transformer, originally introduced by Vaswani et al.~\citep{vaswani2017attention}, has become the backbone of modern models across a wide range of domains, including language, vision, audio, video, graphs, and even time-series data~\citep{achiam2023gpt,team2023internlm,qu2024llama}. Although the Transformer dates back to 2017, its adaptability and robustness have made it indispensable for a variety of tasks. Central to its success is the self-attention mechanism, which is highly effective for sequence modeling, but has quadratic complexity (w.r.t. sequence length), leading to significant computational costs during training. However,
the ability to handle long-context sequences is crucial for large model applications, not only for language tasks but also for multi-modal tasks, where sequences naturally tend to be long~\citep{xue2024longvila}. 
FlashAttention series~\citep{dao2022flashattention,dao2023flashattention,shah2024flashattention} have provided substantial advancements in scaling attention to handle longer sequences by optimizing the CUDA-level computations for better hardware utilization.
However, the theoretical complexity of FlashAttention remains quadratic. Moreover, the need to maintain the KV cache presents further difficulties in managing memory as the sequence length extends~\citep{qin2024various}. As a result, long-sequence processing in Transformer models continues to be a complex and resource-intensive problem.

Recently, numerous variations of attention have been proposed, primarily aimed at addressing its quadratic computational and memory complexity, as well as the growing size of the KV cache~\citep{peng-etal-2023-rwkv,peng2024eagle}. One promising approach line is linear attention~\citep{katharopoulos2020transformers}, which replaces the exponential kernel in softmax attention with a simpler dot product between key and query vectors. This shift allows linear attention to be structured as a linear recurrent neural network (RNN) with matrix-valued hidden states, thereby eliminating the need for a KV cache. In consequence, it supports constant-memory inference and reduces training complexity from quadratic to linear~\citep{yang2023gated}.
A parallel line of research focuses on State Space Models (SSMs), such as Mamba~\citep{gu2023mamba} and Mamba 2~\citep{dao2024transformers}, which draw upon concepts from control theory. Both linear attention and SSMs share a common recurrent formulation, expressed as $\mathbf M_s = {\mathbf M}_{s-1} + \widehat{\mathbf M}_s$, where $\widehat{\mathbf M}_s$ represents the incremental memory state of the $s$-th token~\citep{yang2024parallelizing}.
However, despite these advantages, they tend to perform poorly on recall-intensive tasks, such as in-context learning (e.g., five-shot MMLU~\citep{hendrycks2020measuring}, Phone-book lookup~\citep{jelassi2024repeat}, Needle In A Haystack~\citep{briakou2023searching}) and long-context reasoning. Empirical research~\citep{lieber2024jamba,ren2024samba,waleffe2024empirical,li2025minimax} has shown that models relying solely on linear sequence modeling struggle to excel in these domains. However, a hybrid architecture combining linear sequence modeling layers with standard transformer layers has been demonstrated to significantly enhance model performance on tasks that are recall-intensive.

Sequence Parallelism (SP) techniques~\citep{korthikanti2022reducing,jacobs2023deepspeed,liu2023ring} are commonly employed to partition long sequences into smaller sub-sequences, allowing them to be processed across multiple GPUs in parallel. Despite the advantages offered by SP for handling large sequences, current SP methods do not fully exploit the right-product-first feature of linear attention, which can lead to inefficiencies in parallelism and communication. LASP~\citep{sun2024linear} (referred to as LASP-1) introduced a SP approach specifically tailored for linear attention, that uses a point-to-point (P2P) communication strategy. In this method, intermediate states are transferred across GPUs in a ring-style pattern within the distributed world.
However, although such P2P ring-style communication offers certain benefits, part of its computation has to be executed sequentially, which leads low computation parallelism. In addition, too many small P2P operators make the overlapping of communication and computation difficult.

In this paper, we introduce LASP-2 by rethinking the minimal communication requirement involved in SP of linear attention. Specifically, we innovatively reorganize the whole computation and communication workflow with an optimized execution mechanism. {In this way, only a single all-gather collective communication is needed in the forward or backward of each iteration.} These bring both communication and computation efficiency improvements: 1) the size of intermediate memory state tensor the all-gather operator works on is independent of the sequence length, making the communication burden insignificant in the context of long sequences. The communication parallelism and accessibility to overlap with computation are notably improved. 2) the refactored workflow improves both communication and computation parallelism over multiple devices. Additionally, we separately present LASP-2 with and without masking for autoregressive and bidirectional tasks, respectively, as the presence of a mask matrix significantly impacts the design criterion of LASP-2.
To extend LASP-2 to hybrid models with both linear and standard attention layers, we introduce LASP-2H. This extension employs the same all-gather-based communication for standard attention layers, with a similar designing philosophy on linear attention. We conduct experiments with up to sequence length of 2048K to verify the efficiency advantages of LASP-2 and LASP-2H.

Our main contributions can be summarized as follows:

\begin{compactitem}
\item We rethink the communication design for the current SP on linear attention, reorganize its whole communication \& computation workflow with an optimized execution mechanism. This involves using a single $\texttt{AllGather}$ collective communication on intermediate memory states, whose sizes are independent of sequence length. The resulted LASP-2 improves both communication and computation parallelism for SP on linear attention, thus significantly enhances efficiency.
\item We extend LASP-2 to LASP-2H, offering an efficient SP solution for hybrid models that blend both linear and standard attention layers, employing an unified all-gather-based communication design.
\item We construct a series of Linear-Llama3 models, including both purely linear and hybrid versions. Extensive experimental results on these models with up to a sequence length of 2048K, validate the efficiency improvement and performance of LASP-2 and LASP-2H.
\end{compactitem}

\begin{table*}[t]
    \centering
    \small
    \caption{\textbf{Notations.} Indices, operations, constants, vectors and matrices used in the paper.}
    \label{tab: notation}
    \begin{tabular}{llll}
    \toprule
     $\textbf{Indices}$  &  & $\textbf{Operations}$ &   \\
     $i$  &  Any indices  & $\cdot$ (or omitted) & Matrix multiplication  \\
     $s$  &  Index of current token & $\odot$ & Hadamard multiplication   \\
     $t$  &  Index of chunk & $\textbf{Vectors and Matrices}$ &  \\
     $\textbf{Constants}$ &  & $\mathbf x$, $\mathbf o$ $\in \R^{1\times d}$ & Input and output vectors \\
     $d$ & Hidden dimension & $\mathbf q$, $\mathbf k$, $\mathbf v$ $\in \R^{1\times d}$ & Query, key, value vectors \\
     $W$ & World size & $\mathbf X$, $\mathbf O$ $\in \R^{N\times d}$ & Input and output matrices  \\
     $N$ & Sequence length & $\mathbf Q$, $\mathbf K$, $\mathbf V$ $\in \R^{N\times d}$ & Query, key, value matrices \\
     $T$ & Total number of chunks & $\mathbf M$ $\in \R^{d\times d}$ & Memory state matrix  \\
     $C$ & Chunk length & $\mathbf W_Q$, $\mathbf W_K$, $\mathbf W_V$ $\in \R^{d\times d}$ & Weight matrices \\
    \bottomrule
    \end{tabular}
\end{table*}

\section{Preliminary}
\paragraph{Notation}
In this paper, we ensure the consistent use of notations to enhance clarity. Table~\ref{tab: notation} provides a complete list of all the symbols utilized throughout, including indices, constants, vectors, and matrices. Vectors and matrices are represented in boldface. For simplicity, we have omitted the dimensions related to batch size and number of heads in tensor shapes.

\vspace{-2mm}
\paragraph{Linear Attention}
The term "attention" generally refers to a computation that assigns scores to pairs of positions within a sequence, enabling each element to "attend" to others. The most widely used and significant variant of this mechanism is softmax self-attention, which is central to standard transformer models~\citep{vaswani2017attention}. During training, with an assumption of a single attention head for simplicity, softmax self-attention computes as follows:
\begin{equation}
    \begin{aligned}
        \Q, \K, \V &= \X \mathbf W_Q, \X \mathbf W_K, \X \mathbf W_V, \\
        \mathbf O &= \text{Softmax}(\mathbf Q \mathbf K^\top)\mathbf V.
        \label{eq: attention training}
    \end{aligned}
\end{equation}
The mechanism of pairwise comparisons (induced by materializing $\mathbf Q \mathbf K^\top$) leads to the characteristic quadratic training cost of softmax self-attention.
Recently, Linear Attention~\citep{katharopoulos2020transformers,shen2024scaling,qin2024unlocking} has gained attention as a potential alternative to softmax self-attention, with two key distinctions. First, it removes the $\text{Softmax}(\cdot)$ operation, incorporating it into a kernel feature map. Second, it leverages the associativity of matrix multiplication to reformulate $(\mathbf Q \mathbf K^\top)\mathbf V = \mathbf Q (\mathbf K^\top \mathbf V)$. 
These adjustments reduce both the computation and memory complexity of attention calculation from $O(N^2 d)$ to $O(N d^2)$. This technique is often referred to as the right-product kernel trick because it prioritizes the multiplication on the right side first.

During inference, both softmax self-attention and linear attention handle a single token at each iteration. Given the $s$-th token $\mathbf x_s \in \R^{1\times d}$, softmax self-attention computes requiring the storage of an expanding set of keys $\{\mathbf k_1, \cdots, \mathbf k_s\}$ and values $\{\mathbf v_1, \cdots, \mathbf v_s\}$ i.e., the “KV cache”, which leads to a significant memory burden when dealing with long input sequences.
In linear attention, researchers have experimented with using various nonlinear kernels to replace the $\exp(\cdot)$ function in Eq.~\ref{eq: attention inference}. 
\begin{equation}
\begin{aligned}
    \mathbf q_s, \mathbf k_s, \mathbf v_s &= \mathbf x_s \mathbf W_Q, \mathbf x_s \mathbf W_K, \mathbf x_s \mathbf W_V, \\
    \mathbf o_s &= \frac{\sum_{i=1}^{s} \exp(\mathbf q_s {\mathbf k_i}^\top)\mathbf v_i}{\sum_{i=1}^{s} \exp(\mathbf q_s \mathbf k_i^\top)}.
\end{aligned}
\label{eq: attention inference}
\end{equation}
However, recent studies~\citep{sun2023retentive, yang2023gated, qin2024various} have found that employing a linear kernel (i.e., using the identity function) without a normalizing denominator works effectively in practice.  This results in an unnormalized linear attention form as below: 
\begin{equation}
    \mathbf o_s = \sum_{i=1}^{s} \mathbf q_s ({\mathbf k_i}^\top\mathbf v_i)
     = \mathbf q_s\sum_{i=1}^{s} ({\mathbf k_i}^\top\mathbf v_i)
     = \mathbf q_s \mathbf{M}_s,
     \label{eq: linear attention inference computation}
\end{equation}
where $\mathbf{M}_s = \sum_{i=1}^{s} {\mathbf k_i}^\top\mathbf v_i$ is the prefix sum of ${\mathbf k_i}^\top\mathbf v_i$ from $i=1$ to $s$, which is also known as the memory state in linear attention. This reformulation leads to a recurrent structure for linear attention, resembling the behavior of RNNs as
\begin{equation}
    \mathbf{M}_s =  \mathbf{M}_{s-1} +  \mathbf k_s^\top \mathbf v_s, \quad
    \mathbf o_s = \mathbf q_s \mathbf{M}_s.
\end{equation}

\section{Method}

\subsection{LASP-2 without Masking}
SP methods work by dividing long input sequences into several smaller chunks, which are then distributed across multiple computational devices. Each device independently processes the queries, keys, and values for its assigned chunk in parallel. To complete the attention computation for the entire sequence, necessary communication steps are performed to either gather the results from all devices or exchange information between them. LASP~\citep{sun2024linear} was introduced as a sequence parallelism technique designed specifically for the linear attention module.

Let us consider a distributed computing setup where there are $W$ devices, and the input sequence is divided into $T$ chunks, referred to as the sequence parallel size. In the usual case, $T$ is evenly divisible by $W$, and we often assume $W = T$. It means each chunk is assigned to a single device, ensuring that every chunk is processed in parallel across the distributed system. This scenario exemplifies pure sequence parallelism. Additionally, in Sec.\ref{subsec: hybrid parallel}, we will explore cases where $W \neq T$, representing a hybrid approach that combines sequence parallelism with data parallelism.

In LASP-2, the input sequence $\mathbf X$ is divided into $T$ smaller chunks, represented as $[\mathbf X_t]_{1}^{T}$, and each chunk is distributed across the devices in the distributed system. For each chunk $\mathbf{X}_t$, its corresponding query, key, value, and the linear attention memory state can be computed in parallel across all chunks. This parallel computation is carried out as follows:
\begin{equation}
\begin{aligned}
   \mathbf Q_t, \mathbf K_t, \mathbf V_t &= \mathbf{X}_t \mathbf{W}_Q, \mathbf{X}_t \mathbf{W}_K, \mathbf{X}_t \mathbf{W}_V, \\
   \mathbf{M}_t &= \mathbf{K}_t^\top \mathbf{V}_t.
\end{aligned}
\end{equation}

\begin{algorithm}[t]
\small
    \caption{LASP-2 w/o Masking}
    \label{algo:LASP2 fw without mask}
    \begin{algorithmic}[1]
    \STATE{\textbf{Input:} input sequence $\mathbf X$, distributed world size $W$, sequence parallel size $T=W$.}
    \STATE{Distribute $\mathbf X = [\mathbf X_t]_{1}^{T}$.}
    \FOR{chunk $t \in \{1, \cdots, T\}$ on ranks $\{1, \cdots, W\}$ \textbf{in parallel}}
        \STATE{Calculate $\mathbf Q_t=\mathbf{X}_t \mathbf{W}_Q$, $\mathbf  K_t=\mathbf{X}_t \mathbf{W}_K$, $\mathbf V_t =\mathbf{X}_t \mathbf{W}_V$.}
        \STATE{Compute $\mathbf{M}_{t} = \mathbf K_t^{\top}  \mathbf V_t$.}
        \STATE{Communicate $$~~~~~[\mathbf{M}_t]_{1}^T = \texttt{AllGather}([\mathbf{M}_t]_{1}^T).$$}
        \STATE{Compute $\mathbf{M}_{1:{T}} = \texttt{Sum}([\mathbf{M}_t]_{1}^{T})$.}
        \STATE{Compute $\mathbf{O}_{t} =\mathbf Q_t \mathbf {M}_{1:{T}} $.}
      \ENDFOR
      \STATE{return $\mathbf O = [\mathbf O_t]_{1}^{T}$.}
\end{algorithmic}
\end{algorithm}

By performing this concurrent computation for each chunk, LASP-2 efficiently handles long input sequences in a distributed setting. The query $\mathbf Q_t$, key $\mathbf K_t$, value $\mathbf V_t$, and the memory state $\mathbf M_t$ are calculated individually for every chunk of the sequence, ensuring that no single device is overburdened with processing the entire sequence at once. This distributed approach facilitates better memory management and computational efficiency, especially when dealing with extremely long sequences. Thus, LASP-2 leverages the power of sequence partitioning to optimize the calculation of linear attention in a distributed framework.

Notably, in LASP-2, only a single all-gather collective communication operation is required during the forward pass. This all-gather operation acts on the memory states $[\mathbf M_t]_1^T$ associated with each sequence chunk, ensuring that every device in the system has access to the complete set of memory states $[\mathbf M_t]_1^T$. Once the memory states from all chunks have been gathered, they are concurrently accumulated on all devices to compute the memory state corresponding to the entire input sequence. This process is expressed as follows:
\begin{equation}
\begin{aligned}
   [\mathbf{M}_t]_{1}^T = \texttt{AllGather}([\mathbf{M}_t]_{1}^T), \\
   \mathbf{M}_{1:{T}} = \texttt{Sum}([\mathbf{M}_t]_{1}^{T}).
\end{aligned}
\end{equation}
Finally, the linear attention output corresponding to the local query $\mathbf{Q}_t$ can be computed as:
\[
\mathbf{O}_{t} = \mathbf Q_t \mathbf {M}_{1:{T}}.
\]

Importantly, the accumulation step $\texttt{Sum}([\mathbf{M}_t]_{1}^{T})$ can be efficiently performed in a recursive manner, by adding each memory state sequentially as $\mathbf{M}_{1:{t-1}} + \mathbf{M}_t$. This eliminates the need to repeatedly calculate the sum of the memory states from earlier chunks, improving the efficiency of the computation. To further optimize performance, we cache the accumulated result $\mathbf{M}_{1:{T}}$ in high-bandwidth memory (HBM). This caching strategy speeds up the backward pass by avoiding redundant recalculations of $\mathbf{M}_{1:{T}}$, which is necessary for computing gradients. This approach is akin to the concept of activation checkpointing, where intermediate activations are saved to avoid recomputation.

It is important to point out that each memory state $\mathbf{M}_t$ has dimensions of $d \times d$, which means the communication cost for the all-gather operation is independent of the sequence or chunk length. Instead, the cost scales linearly with the number of devices involved in the SP communication group. 
For clarity, we provide a summary of the LASP-2 method, without considering the attention mask, in Algorithm~\ref{algo:LASP2 fw without mask}. During the backward pass, a similar all-gather communication operation on the gradients of memory states $\mathbf{dM}_t$ is required. The details of this backward pass without masking, can be found in Algorithm~\ref{algo:LASP2 bw without mask} in Appendix~\ref{app: backward algo} for further reference.

\subsection{LASP-2 with Masking}

\begin{algorithm}[t]
\small
    \caption{LASP-2 w/ Masking}
    \label{algo:LASP2 fw with mask}
    \begin{algorithmic}[1]
    \STATE{\textbf{Input:} input sequence $\mathbf X$, distributed world size $W$, sequence parallel size $T=W$.}
    \STATE{Distribute $\mathbf X = [\mathbf X_t]_{1}^{T}$.}
     \STATE{Initialize mask matrix $\mathbf \Psi$, where $\mathbf \Psi_{ij} = 1$ if $i \geq j$ and $\mathbf \Psi_{ij} = -\infty$ if $i < j$.}
    \FOR{chunk $t \in \{1, \cdots, T\}$ on ranks $\{1, \cdots, W\}$ \textbf{in parallel}}
        \STATE{Calculate $\mathbf Q_t=\mathbf{X}_t \mathbf{W}_Q$, $\mathbf  K_t=\mathbf{X}_t \mathbf{W}_K$, $\mathbf V_t =\mathbf{X}_t \mathbf{W}_V$.}
        \STATE{Compute $\mathbf{M}_{t} = (\mathbf K_t)^{\top}  \mathbf V_t$.}
        \STATE{\color{magenta}{Communicate $[\mathbf{M}_t]_{1}^T = \texttt{AllGather}([\mathbf{M}_t]_{1}^T)$.}}
        \STATE{\color{cyan}{Compute $\mathbf O_{\mathrm{t, intra}}= [(\mathbf Q_t \mathbf K_t^{\top }) \odot \mathbf \Psi]\mathbf V_t$.}}
        \STATE{Compute prefix sum $\mathbf{M}_{1:{t-1}} = \texttt{PrefixSum}([\mathbf{M}_t]_{1}^{t-1})$.}
        \STATE{Compute $\mathbf{O}_{\mathrm{t, inter}} = \mathbf Q_t \mathbf{M}_{1:{t-1}}$.}
        \STATE{Compute $\mathbf O_t=\mathbf O_{\mathrm{t, intra}}+ \mathbf{O}_{\mathrm{t, inter}}$.}
      \ENDFOR
      \STATE{return $\mathbf O = [\mathbf O_t]_{1}^{T}$.}
\end{algorithmic}
\end{algorithm}

\begin{figure}[t]
    \centering
    \includegraphics[width=0.55\columnwidth]{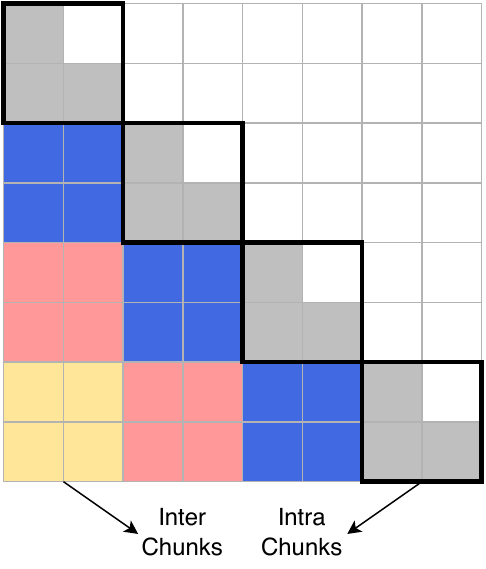}
    \caption{\textbf{Computation Decomposition in LASP-2 with masking.} Colored chunks represent inter-chunks.}
    \vspace{-4mm}
    \label{fig: computation decomposition}
\end{figure}

In autoregressive tasks, the mask matrix $\mathbf \Psi \in \{-\infty, 1\}^{N \times N}$ is typically a lower triangular matrix, where $\mathbf \Psi_{ij} = 1$ for $i \geq j$ and $\mathbf \Psi_{ij} = -\infty$ when $i < j$. This structure enforces a causal constraint during computation. Specifically, when calculating $\mathbf O = \text{Softmax}(\mathbf Q \mathbf K^\top \odot \mathbf \Psi)\mathbf V$, it becomes impossible to leverage the associative property of matrix multiplication to reduce the computational complexity from quadratic to linear in a parallel form.

To address this challenge in linear attention with a causal mask, we adopt the approach of computation decomposition, as proposed in earlier work~\citep{yang2023gated,sun2024linear}. Figure~\ref{fig: computation decomposition} provides an illustration that highlights the difference between intra-chunk and inter-chunk computations in linear attention. Inter-chunk calculations, which have no dependencies on other chunks across devices, can be treated as if they have no causal mask. As a result, these computations can be parallelized across all devices in the distributed setup. In contrast, intra-chunk calculations account for the influence of previous chunks ($1$ to $(t-1)$) on the $t$-th chunk. These intra-chunk operations are affected by the mask matrix, and therefore, require specialized handling to respect the causal constraints.



For linear attention computation on intra-chunks, given the query, key, and value matrices $\mathbf{Q}_t$, $\mathbf{K}_t$, and $\mathbf{V}_t$ corresponding to the chunk $\mathbf{X}_t$, the output is computed as
\begin{equation}
   \mathbf{O}_{\mathrm{t, intra}} = [(\mathbf{Q}_{t} \mathbf{K}_{t}^\top) \odot \mathbf{\Psi}] \mathbf{V}_{t},
\end{equation}
This formulation adheres to the standard left-product matrix multiplication. Although the computation can be executed in parallel across devices, it retains the quadratic complexity commonly associated with traditional attention mechanisms during training. This limitation arises from the element-wise masking operation ($\odot \mathbf{\Psi}$), which enforces causal constraints within the chunk, preventing the use of optimizations that would reduce the computational cost to linear.

For linear attention computation across inter-chunks, we follow a similar approach as the procedure outlined for LASP-2 without masking. First, the memory states for each chunk are computed concurrently across different devices as $\mathbf{M}_t = \mathbf{K}_t^\top \mathbf{V}_t$. These memory states, corresponding to each chunk, are initially distributed across separate devices. To synchronize the results, an $\texttt{AllGather}$ collective communication operation is performed. This step ensures that all devices hold the memory states for all chunks, enabling further parallel processing.
Once the memory states have been gathered, we proceed with a concurrent $\texttt{PrefixSum}$ operation across all devices. This operation accumulates the memory states from the 1st chunk up to the $(t-1)$-th chunk, effectively building the necessary intermediate states. This can be expressed as:
\begin{equation}
\begin{aligned}
    \relax [\mathbf{M}_t]_{1}^T &= \texttt{AllGather}([\mathbf{M}_t]_{1}^T), \\
    \mathbf{M}_{1:{t-1}} &= \texttt{PrefixSum}([\mathbf{M}_t]_{1}^{t-1}).
\end{aligned}
\end{equation}

The $\texttt{PrefixSum}$ operation can be optimized by implementing it recursively, utilizing cached memory states stored on the HBM. Specifically, the accumulation of memory states is computed as:
\begin{equation}
    \mathbf{M}_{1:{t-1}} = \mathbf{M}_{1:{t-2}} + \mathbf{M}_{t-1}.
\end{equation}
By caching $\mathbf{M}_{1:{t-1}}$, the backward pass computation is facilitated since this cached value is a necessary activation for gradient calculations. This approach not only speeds up the backward pass but also reduces the computational load, as the cached memory state eliminates the need for repeated re-computation.

Following the calculation of the memory states, the outputs corresponding to the inter-chunks and the final output for the $t$-th token can be derived with ease. The overall output for the $t$-th token is obtained by summing both the intra-chunk and inter-chunk outputs.
\begin{equation}
    \mathbf{O}_{\mathrm{t, inter}} = \mathbf Q_t \mathbf{M}_{1:{t-1}}, \quad
    \mathbf O_{t} = \mathbf{O}_{t,\mathrm{intra}} + \mathbf{O}_{t,\mathrm{inter}}.
\end{equation}

We provide the complete algorithm for LASP-2 with masking in Algorithm~\ref{algo:LASP2 fw with mask}, and its backward pass in Algorithm~\ref{algo:LASP2 bw with mask} in Appendix~\ref{app: backward algo}. Note that, in Algorithm~\ref{algo:LASP2 fw with mask}, the communication operation in line 7 (in magenta), along with the computation of $\mathbf O_{\mathrm{t, intra}}$ in line 8 (in cyan), can be overlapped by executing them on separate threads. This concurrent execution helps improve overall efficiency, as it allows for the overlap of communication and computation.

\subsection{LASP-1 vs LASP-2}

LASP-2, as well as its previous version LASP-1, both aim on efficient SP on linear attention. Although, in theory, LASP-1 and LASP-2 share similarity on communicating the KV activation ($d \times d$), whose size is independent of the sequence or chunk length. They have fundamental distinctions where the key differences lie in their communication manners and the computational order reorganization, as elaborated as below:

\begin{compactitem}
\item LASP-1 utilizes a ring-style P2P communication, which needs to launch many \texttt{send} \& \texttt{receive} operators between devices, to sequentially transfer the KV activation one-by-one among the devices. This makes the communication process relatively slow and hard to adequately overlap with intra-chunk computations.
\item While LASP-2 uses a single \texttt{AllGather} collective communication operator to exchange KV activation concurrently among all decices. This offers practical advantages: (1) Only one well-optimized collective communication operator needs to be launched, and the exchange of KV activation on all devices can be finished concurrently all at once; (2) the collective communication can be more easily overlapped with computations. Like in LASP-2 with masking, the \texttt{AllGather} communication is able to overlap with the intra-chunk output computations. And, in addition, LASP-2 reorganizes the whole computation order to make the \texttt{AllGather} based communication strategy feasible and efficiency. 
\end{compactitem}

We also write down the Algorithms of LASP-1 (with and without masking) in identical mathematical symbols in Appendix~\ref{app: lasp-1 algo} for convenience to compare with LASP-2 on their algorithmic differences.


\begin{figure*}[t]
    \centering
    \includegraphics[width=0.9\textwidth]{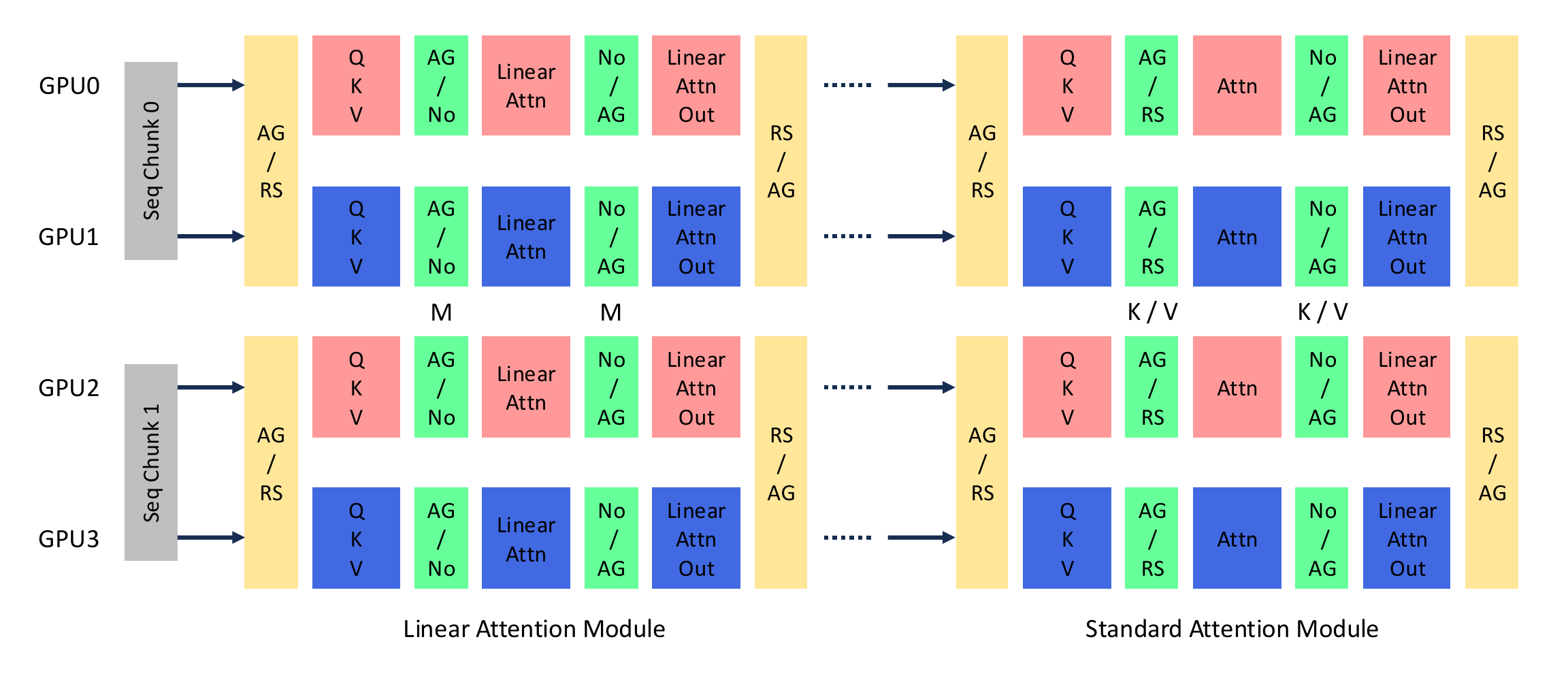}
    \vspace{-6mm}
    \caption{\textbf{Visualization of LASP-2H on Linear Attention and Standard Attention hybrid model.} We exemplify LASP-2H on the hybrid layers of linear attention and standard attention modules with both TP and SP (both have a dimension of 2). The communication operations colored in yellow and green are for TP and SP, respectively. AG/RS: all-gather in forward and reduce-scatter in backward, and vice versa. AG/No: all-gather in forward and no-op in backward, and vice versa. Note that the SP communication operations for linear attention operate on the memory state $\mathbf{M}_t\in \R^{d\times d}$, while for standard attention, they operate on states $\mathbf{K}_t, \mathbf{V}_t\in \R^{C\times d}$.}
    \vspace{-4mm}
    \label{fig: lasp2h visualization}
\end{figure*}

\subsection{Theoretical Cost Analysis}
For better understanding the superiorities of LASP-2, we provide a theoretical cost analysis of both LASP-1 and LASP-2. We consider the pure SP scenario, i.e., the distributed world size is $W$, and an input sequence with a length of $N$ is partitioned into $T=W$ chunks, thus all devices in this world need to involve into the communication. Below $B$ denotes batch size, $H$ represents number of heads.

\textbf{Communication traffic in each communication step: LASP-1: $BHd^2$, LASP-2: $BHd^2$.} This is because both LASP-1 and LASP-2 transfer linear attention memory states (not keys and values) among devices. The memory state corresponding to each chunk (located at each device) has a tensor shape of [B, H, d, d]. Thus in each communication step, their communication traffic are both $BHd^2$. 

For a Linear-Llama3-1B model with $B=16$, $H=16$ and $d=2048$, each memory state will has $BHd^2 \approx 1.07$B parameters, which takes around $2.14 GB$ memory in FP16. For a Linear-Llama3-8B model with $B=16$, $H=32$ and $d=4096$, each memory state has $BHd^2 \approx 8.59$B parameters, which takes around $17.18 GB$ memory in FP16.

\textbf{Number of communication steps in each iteration: LASP-1: $2(W-1)$, LASP-2: $2$.} This depends on the different communication manners of these two algorithms. During the forward of an iteration, LASP-2 launches a single all-gather operation to gather all memory states $\mathbf M_t$ to all devices, i.e., $[\mathbf M_t]_1^T = \texttt{AllGather}([\mathbf M_t]_1^T)$. This collective operation is concurrently executed on all devices. While in backward, another all-gather is performed on the gradients of $\mathbf M_t$, i.e., $[\mathbf {dM}_t]_1^T = \texttt{AllGather}([\mathbf {dM}_t]_1^T)$. Thus in each iteration, LASP-2 has 2 communication steps. While LASP-1 uses a pair of {send} \& {receive} operation to sequentially exchange the memory state from one device to another device. During forward, device $i$ sends its memory state to device $i+1$, and device $i+1$ receives the memory state from device $i$, and so on. Computations of $\mathbf O_{\mathrm{t, inter}}$, $\mathbf O_t$ and updates of $\mathbf M_{t}$ are followed behind each {receive} operation on that device. Thus in the process of forward, LASP-1 has $W-1$ communication steps. In the backward, this process is repeated reversely from the last device to device 0. Thus in each iteration, LASP-1 have totally $2(W-1)$ communication steps.

Given that both LASP-1 and LASP-2 perform a total of $I$ iterations, their communication traffic models can be expressed as follows: LASP-1: $2(W-1)IBHd^2$ and LASP-2: $2IBHd^2$. Ideally, the communication traffic of LASP-2 would be reduced by a factor of $W-1$ compared to LASP-1. However, the actual communication cost depends on practical factors like communication bandwidth, which is typically faster within nodes and slower across nodes, and communication stability. As a result, the benefits of LASP-2 become more evident in clusters with slower interconnects, and vice versa. It is important to note that this cost model only accounts for communication, excluding computation or data-loading. In practice, communication represents a smaller portion of the total cost, thus the overall training speedup achieved by LASP-2 is less than $W-1$ times. LASP-2 performs best in scenarios involving long sequences, large clusters, slow communication links, and efficient data-loading and computation.

\subsection{Hybrid Model Sequence Parallelism}

The hybrid model, which combines linear transformer layers with standard transformer layers that utilize softmax self-attention, has been demonstrated to effectively enhance long-context capabilities, particularly in tasks such as recall and retrieval. To optimize SP in such hybrid models, we propose an extended version of LASP-2, referred to as LASP-2H. This approach introduces a comprehensive solution by incorporating SP into both the linear attention and standard attention modules. The structure of LASP-2H is illustrated in Fig.~\ref{fig: lasp2h visualization}.

\textbf{On Linear Attention Module.} As outlined in Algorithm~\ref{algo:LASP2 fw without mask} and Algorithm~\ref{algo:LASP2 fw with mask}, LASP-2H handles linear attention modules by performing a single all-gather communication operation on the memory state $\mathbf{M}_t \in \R^{d \times d}$. 
The communication complexity remains independent of both sequence or chunk length, and only scales linearly with the SP size $T$, making this method efficient in distributed clusters.

\textbf{On Standard Attention Module.} 
Context Parallelism (CP) is a SP technique in Megatron-LM that divides network inputs and all activations along the sequence dimension. This approach is specifically tailored for standard softmax attention. While traditional CP implementations in Megatron-LM rely on overlapping communication and computation in a ring-like structure~\citep{liu2023ring}, our LASP-2H adopts a different method, following the best practice in Llama3~\citep{dubey2024llama}. Instead of the ring-style strategy, LASP-2H employs AllGather-based communication on standard attention, where the $\mathbf K_t$ and $\mathbf V_t$ tensors are first gathered across devices, after which the attention output is computed locally for the $\mathbf Q_t$ tensor chunk.
Although the all-gather communication has a higher latency compared to ring-based methods, it provides greater ease and flexibility in handling various types of attention masks, such as document-level masks. This flexibility is particularly beneficial in scenarios where different attention patterns are needed. Additionally, the all-gather latency is minimized because the $\mathbf K_t$ and $\mathbf V_t$ tensors are significantly smaller than the $\mathbf Q_t$ tensor, especially when using Grouped Query Attention (GQA)~\citep{ainslie2023gqa}. As a result, the time complexity of computing the attention output far exceeds the complexity of all-gather operation. We present the description of AllGather-based Context Parallelism in Algorithm~\ref{algo:allgather_context_parallelism} in Appendix~\ref{app: AllGather-based Context Parallelism}.

\section{Experiments}
We conducted an empirical evaluation of LASP-2 by applying it to a model based on Llama3~\citep{dubey2024llama}. We replaced the standard softmax attention with various linear attention modules, including the original basic linear attention~\citep{katharopoulos2020transformers}, Lightning Attention~\citep{qin2024lightning}, Retention~\citep{sun2023retentive}, Gated Linear Attention (GLA)~\citep{yang2023gated}, Based~\citep{arora2024simple}, and Rebased~\citep{aksenov2024linear}. This modified model, termed Linear-Llama3, comprises 16 (linear transformer) layers, with a total of 1B parameters.
Additionally, we created a hybrid model by retaining transformer layers with standard softmax attention at every fourth layer of Linear-Llama3, forming a $1/4$ hybrid architecture. All experiments were conducted on the SlimPajama dataset~\citep{cerebras2023slimpajama}, utilizing the Llama3 tokenizer~\citep{dubey2024llama}. The full dataset contains 627B tokens, but for our experiments, we used a 50B tokens subset derived from the first chunk of the training split. 
The experiments were performed using GPT-style autoregressive language modeling tasks with attention masks, as this setup mirrors many practical scenarios where such tasks are commonly applied. Note that the primary focus of these experiments is to assess the training efficiency of LASP-2 when handling very-long input sequences. Training a large language model with optimal long-context capabilities falls outside the scope of this study.

Besides the following results, we have provided more additional experiment results in Appendix~\ref{app: additional exp}.

\subsection{Experimental Setup}

\textbf{Hardware and Software.} Our experiments were conducted on a configuration of up to 16 DGX-A100 servers, each equipped with 8 A100 GPUs. The GPUs are connected through NVSwitch, offering an inter-GPU bandwidth of 600 GBps. The experiments were implemented using PyTorch 2.3.1, with support from CUDA 12.1, cuDNN 8.9.2, and NCCL 2.20.5. The algorithm was developed on top of NVIDIA's Megatron-Core 0.9.0~\citep{shoeybi2019megatron}. We use Triton 2.3.1~\citep{Tillet2019TritonAI} to accelerate the linear attention computation on GPU, and take FlashAttention-2~\citep{dao2023flashattention} as the standard attention implementation. When implement other SP methods (e.g., Ring Attentoin, Megatron-SP) on linear attention instances for the purpose of comparison, we do not incorporate the right-product kernel trick. We maintain the use of each method's original communication primitives and computational manners as they originally proposed for standard attention.

\textbf{Hyperparameters.} For training the Linear-Llama3 model, we employed a cosine learning rate schedule with a linear warm-up phase~\citep{sun2024co2}. The minimum learning rate was set to $1e^{-6}$. We applied gradient clipping with a value of 1.0 and weight decay at a rate of 0.1. The Adam optimizer~\citep{kingma2014adam} was used, configured with $\beta_1=0.9$ and $\beta_2=0.95$ \citep{zhang2019fast,zhou2020pbsgd}. Additionally, the dropout rate in both attention and hidden layers was set to 0~\citep{tang2023ms}.

\begin{table*}[t]
    \centering
    \small
    \caption{\textbf{Convergence Performance Results.} All experiments used 8 A100 GPUs, sequence length of 16K, and batch size of 8, trained on 50B tokens from the SlimPajama corpus.}
    \vspace{-2mm}
        \begin{tabular}{lllcccc}
        \toprule
            \multirow{2}{*}{\textbf{Model}}   & \multirow{2}{*}{\textbf{SP Method}}   & \multirow{2}{*}{\textbf{Attention Module}}    & \multicolumn{2}{c}{\textbf{Pure Model}}        &  \multicolumn{2}{c}{\textbf{$1/4$ Hybrid Model}}  \\
            \cmidrule(ll){4-5} \cmidrule(ll){6-7}
                   &  &  & \textbf{Thpt}   & \textbf{Loss}     & \textbf{Thpt}     &  \textbf{Loss}  \\  \midrule
            Llama3 & Ring Attention & Standard Attention & 16549.5 & 2.759 & $\backslash$ & $\backslash$ \\ \midrule
            \multirow{6}{*}{\makecell[c]{Linear-Llama3}} & \multirow{6}{*}{\makecell[c]{LASP-2(H)}} &
            Basic Linear Attention        &     17834.3     &  2.892 & 17394.7 &  2.824 \\ 
            &    &  Lightning Attention   &  17926.1  & 2.862 & 17384.2 &  2.758  \\ 
            &   &  Retention   &  17859.6  & 2.867 & 17352.5 &  2.759  \\ 
            &   &   GLA        &     17785.3   &  2.845   & 17273.2 & 2.754 \\ 
            &   &  Based       &    17946.1   &  2.754  & 17462.5 & 2.751 \\ 
            &   &   Rebased       &   17896.2    & 2.845  & 17284.5 & 2.787 \\
        \bottomrule             
        \end{tabular}
 \label{tab:performance}
\vspace{-4mm}
\end{table*}

\begin{figure}[t]
    \centering
    \includegraphics[width=0.49\textwidth]{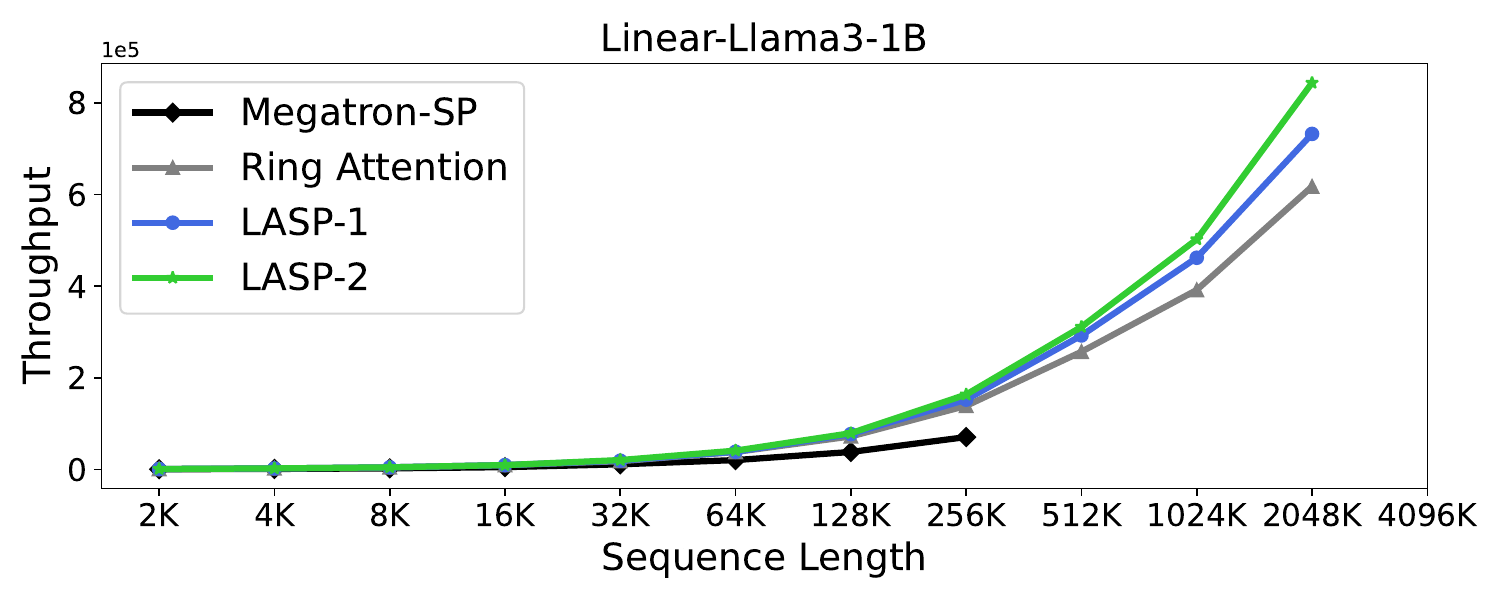}
    \caption{\textbf{Speed Comparison (tokens/s).} Experiments were carried out on a pure Linear-Llama3-1B model, utilizing the basic linear attention module. A total of 64 A100 GPUs were employed, and the SP size $T$ was also set to 64. To accommodate very-long sequence lengths, such as 2048K, the batch size was kept fixed at 1 throughout this experiment.}
    \label{fig: lasp2_speed}
    \vspace{-4mm}
\end{figure}

\begin{figure}[t]
    \centering
    \includegraphics[width=0.49\textwidth]{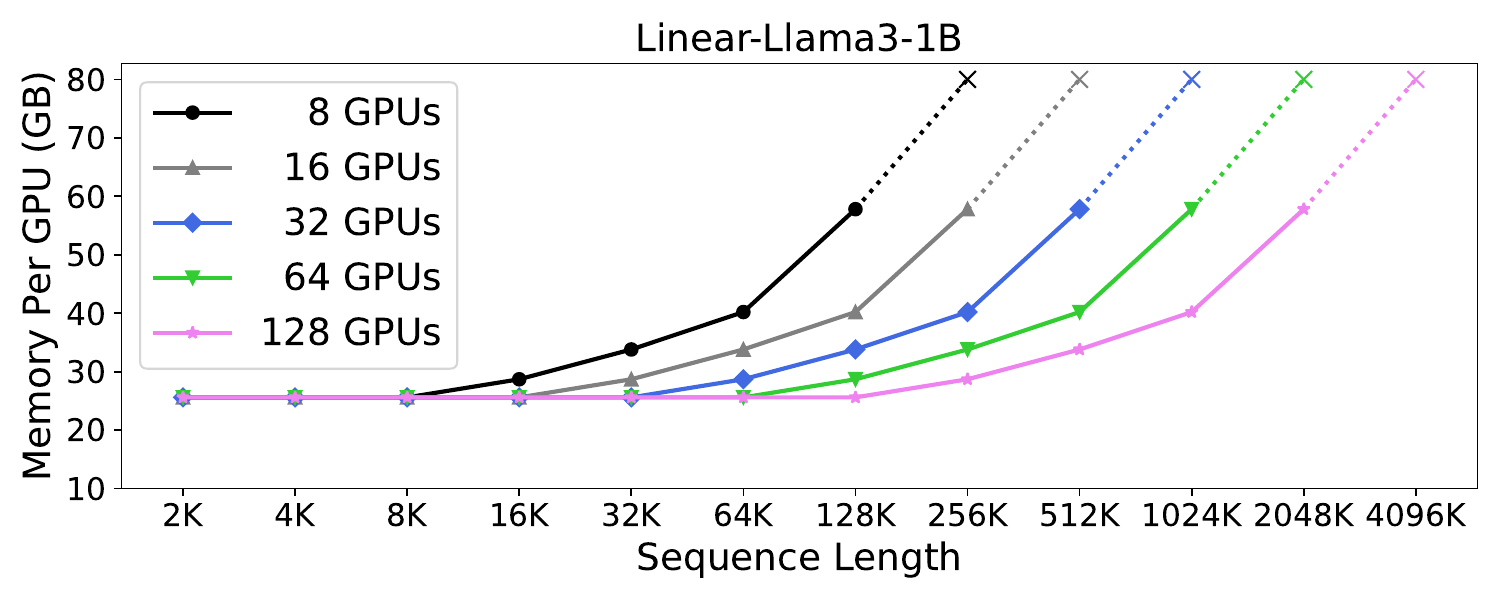}
    \includegraphics[width=0.49\textwidth]{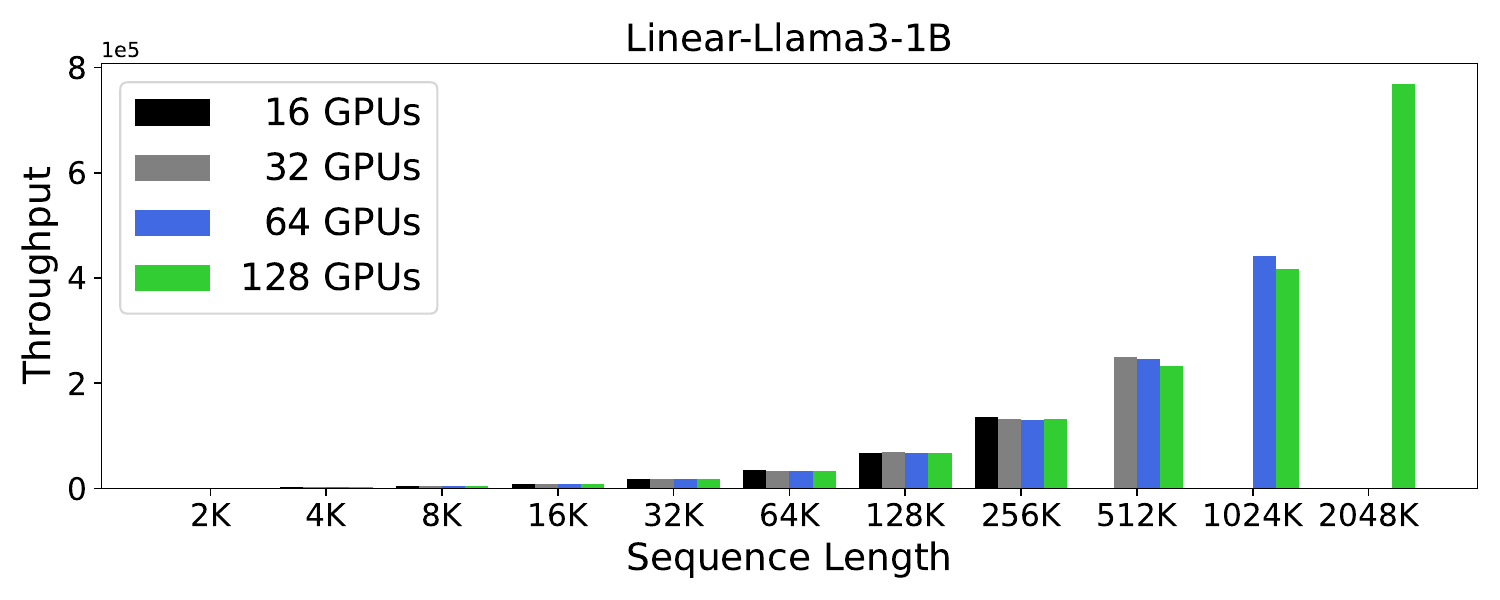}
    \caption{\textbf{Scalability Results.} Experiments were conducted on a pure Linear-Llama3-1B model using the Basic Linear Attention module. SP size $T$ was always equal to number of GPUs. Batch size was fixed as 1 to accommodate very-long sequence lengths, e.g., 2048K. The sign "$\times$" with a dotted line represented occurring an Out of Memory (OOM).}
    \label{fig: lasp2_scalability}
    \vspace{-5mm}
\end{figure}

\subsection{Speed}
To assess the speed performance of our proposed LASP-2, we conducted a comparison against existing SP methods, including Megatron-SP~\citep{korthikanti2022reducing}, Ring Attention~\citep{liu2023ring}, and LASP-1~\citep{sun2024linear}. As depicted in Fig.~\ref{fig: lasp2_speed}, LASP-2 demonstrated superior throughput, particularly when sequence lengths exceeded 64K. This performance advantage became increasingly prominent as sequence lengths grew longer. Specifically, at a sequence length of 512K, LASP-2 outperformed Ring Attention by 17.8\% and surpassed LASP-1 by 7.3\%. This advantage became even more pronounced at a sequence length of 2048K, where LASP-2 achieved throughput gains of 36.6\% over Ring Attention and 15.2\% over LASP-1.

\subsection{Scalability}

We assessed the scalability of LASP-2 in terms of both GPU memory usage and throughput by adjusting the sequence length and the number of GPUs. The results were displayed in Figure \ref{fig: lasp2_scalability}. LASP-2 demonstrated the ability to scale linearly with the input sequence length by increasing the number of GPUs. For instance, while maintaining the same memory cost per GPU, using 8 GPUs allowed training on sequences up to 128K in length, whereas 128 GPUs (16 $\times$ 8 GPUs) enabled training on sequences as long as 2048K (16 $\times$ 128K). Additionally, we observed that increasing both sequence length and device numbers results in higher throughput, indicating improved communication efficiency and linear scalability. More detailed quantitative scalability outcomes are provided in Table \ref{tab:quantitative} in Appendix \ref{app: additional exp}.

\subsection{Convergence Performance}

We conducted additional experiments to assess the pretraining convergence performance of LASP-2 on Llama-3 with various attention modules, including standard softmax attention, basic linear attention, Lightning Attention, Retention, GLA, Based, Rebased, and their $1/4$ hybrid models. All experiments were performed on the SlimPajama corpus \citep{cerebras2023slimpajama}, using 50B tokens, a sequence length of 16K, and a global batch size of 8, using 8 A100 GPUs. The results, as shown in Table \ref{tab:performance}, indicated that for pure Linear-Llama3 models with different linear attention modules, LASP-2 achieved comparable, though slightly higher, loss values while maintaining superior throughput. On the $1/4$ hybrid Linear-Llama3 model, the loss results were generally better than those of the pure linear models, with Lightning Attention, Retention, and GLA even attaining equivalent or lower loss values compared to the baseline. The Based attention module shows strong throughput and loss performance, since its original design uses a mix of (Taylor) linear attention and sliding window attention.
The $1/4$ hybrid model striked a balance between throughput and convergence performance, performing competitively when compared to both the baseline and its pure linear version.

\subsection{Related Work}

\subsubsection{Linear Sequence Modeling}

\textbf{Linear Attention.} Vanilla linear attention~\citep{katharopoulos2020transformers} introduces the use of kernel methods as a replacement for the $\operatorname{Softmax}$ attention~\citep{vaswani2017attention}, thereby reducing the computational complexity to linear in sequence length. Following this, several variants of linear attention have been proposed. TransNormerLLM~\citep{qin2023scaling,qin2023transnormerllm} proposes Lightning Attention, a refined linear attention mechanism that accelerates processing by optimizing IO interactions. Lightning Attention-2~\citep{qin2024lightning} further realizes the theoretical advantages of linear attention by separately handling inter- and intra-block computations. RetNet~\citep{sun2023retentive} introduces a retention mechanism that combines recurrence with attention, benefiting from both parallel training and linear inference. Gated Linear Attention (GLA)~\citep{yang2023gated} incorporates a data-independent gating mechanism into the linear attention framework, and presents an efficient algorithm for training. DeltaNet~\citep{Schlag2021LinearTA} and its parallelized version~\citep{yang2024parallelizing} use a delta rule-like update to enhance linear attention performance in long-context scenarios. Finally, Gated Slot Attention (GSA)~\citep{zhang2024gsa}, inspired by GLA, introduces a gated linear attention mechanism with bounded-memory slot control to further improve efficiency.

\textbf{State Space Modeling.} The SSM serves as a powerful framework for representing the behavior of sequences within dynamic systems, and it has shown considerable promise in the realm of linear sequence modeling. Mamba \citep{gu2023mamba} incorporates a mechanism for selecting states, thereby facilitating the scaling of linear sequence lengths. This architecture has been further enhanced in Mamba-2 \citep{dao2024transformers}, where the introduction of the state space duality (SSD) framework optimizes its performance.

\textbf{Linear RNN.} Traditional RNNs face significant challenges in handling long-context sequence modeling, primarily due to their inherent sequence dependency during training, which prevents them from fully capitalizing on scaling laws~\citep{sun2023retentive}. To address these limitations, RWKV~\citep{peng-etal-2023-rwkv,peng2024eagle} was introduced as a linear RNN-based large language model that aims to efficiently manage long-term dependencies. Additionally, HGRN~\citep{qin2024hierarchically} highlights the critical role of data-dependent decay mechanisms in enhancing linear RNN performance, demonstrating how adjustments to decay parameters can improve learning in long-context tasks. An enhanced version, HGRN2~\citep{qin2024hgrn2}, expands on this approach by incorporating a state expansion mechanism that utilizes outer product operations, which allows for greater scalability and improved modeling capabilities over extended sequences. Both RWKV and HGRN series seek to overcome weaknesses of RNNs for efficient long-sequence modeling.

\subsubsection{Sequence Parallelism}
SP \citep{li2022sequence} is a distributed technology designed for training language models more efficiently, which is implemented by dividing a long sequence into multiple shorter subsequences and processing these subsequences in parallel on multiple computing devices. Existing SP methods \citep{korthikanti2022reducing,jacobs2023deepspeed} whose parallelism degree cannot exceed the number of attention heads, which limits their scalability. Ring Attention \citep{liu2023ring} is proposed to address high memory cost in long sequence modeling by distributing subsequences across different devices and overlapping the communication of KV blocks. LASP \citep{sun2024linear} proposes a new linear attention-tailored SP strategy based on GPU friendly implementation by utilizing a P2P ring-style communication strategy, but still lacks of optimizations for hybrid model architecture.

\section{Conclusion}

This paper presents LASP-2, a new SP method that addresses the inefficiencies of existing SP approaches for linear sequence modeling. By redesigning the whole algorithm workflow and leveraging a single all-gather communication strategy, LASP-2 significantly enhances both the communication and computation parallelism, and enables easier communication-computation overlapping, comparing with preceding work LASP-1. Our results demonstrate that LASP-2 offers significant improvements in speed and scalability, especially in the context of very-long sequence length. Furthermore, the extension to LASP-2H enables efficient SP in hybrid models that integrate both linear and standard attention modules, both utilize an unified all-gather-based communication primitive. Experimental evaluations on the Linear-Llama3 models validate these advancements, with LASP-2 outperforming previous methods like LASP-1 and Ring Attention by substantial margins, particularly at extreme sequence lengths. These findings confirm the practical utility of LASP-2 for large-scale distributed systems, making it a promising approach for future applications in long-sequence linear transformer models.




\newpage
\section*{Impact Statement}

This work represents a notable advancement in artificial intelligence and machine learning, particularly in improving the efficiency and scalability of linear attention-based models. LASP-2 enables the processing of much longer sequences compared to existing methods while significantly accelerating computation, making it highly beneficial for tasks like natural language understanding, genomic sequence analysis, and time-series forecasting. However, the enhanced capabilities and efficiency introduced by LASP-2 also raise ethical and societal considerations, such as the potential for misuse in generating persuasive but misleading content or in surveillance applications. Nevertheless, the contributions of LASP-2 to reducing computational overhead and energy consumption in training large models may also bring positive environmental impacts.

\nocite{langley00}

\bibliography{example_paper}
\bibliographystyle{icml2025}

\appendix
\onecolumn
\section{Appendix}
\label{appendix}

\subsection{LASP-2 Algorithms (Backward Pass)}
\label{app: backward algo}

See Algorithm~\ref{algo:LASP2 bw without mask} and Algorithm~\ref{algo:LASP2 bw with mask}.

\begin{algorithm*}[ht]
\small
    \caption{LASP-2 w/o Masking (Backward Pass)}
    \label{algo:LASP2 bw without mask}
    \begin{algorithmic}[1]
    \STATE{\textbf{Input:} distributed world size $W$, sequence parallel size $T=W$, $\mathbf Q_t,\mathbf K_t,\mathbf V_t,\mathbf O_t,\mathbf{dO}_t \in \mathbb{R}^{C \times d}$ for chunk $t\in \{1, \cdots, {T}\}$.}
    \FOR{chunk $t\in \{1, \cdots, T\}$ on ranks $\{1, \cdots, W\}$ \textbf{in parallel}}
        \STATE{Compute $\mathbf{dM}_{t} = (\mathbf Q_t)^{\top}  \mathbf {dO}_t$.}
        \STATE{Communicate $[\mathbf{dM}]_{1}^T = \texttt{AllGather}([\mathbf{dM}]_{1}^T)$.}
        \STATE{Compute $\mathbf{dM}_{1:T} = \texttt{Sum}([\mathbf{dM}]_{t+1}^T)$.}
        \STATE{Compute
        $\mathbf{dQ}_{\mathrm{t}}=\mathbf{dO}_t \mathbf{M}_{1:T}^{\top}$.}
        \STATE{Compute
        $\mathbf{dK_{\mathrm{t}}}={\mathbf V_t}\mathbf{dM}_{1:T}^{\top}$.}
        \STATE{Compute
        $\mathbf{dV_{\mathrm{t}}}=\mathbf K_t \mathbf{dM}_{1:T}$.}
    \ENDFOR
    \STATE{return $\mathbf {dQ}=[\mathbf {dQ}_t]_1^T$, $\mathbf{dK}=[\mathbf{dK}_t]_1^T$, $\mathbf{dV}=[\mathbf{dV}_t]_1^T$.}
\end{algorithmic}
\end{algorithm*}

\begin{algorithm*}[ht]
\small
    \caption{LASP-2 w/ Masking (Backward Pass)}
    \label{algo:LASP2 bw with mask}
    \begin{algorithmic}[1]
    \STATE{\textbf{Input:} distributed world size $W$, sequence parallel size $T=W$, $\mathbf Q_t,\mathbf K_t,\mathbf V_t,\mathbf O_t,\mathbf{dO}_t \in \mathbb{R}^{C \times d}$ for chunk $t\in \{1, \cdots, {T}\}$.}
    \FOR{chunk $t\in \{1, \cdots, T\}$ on ranks $\{1, \cdots, W\}$ \textbf{in parallel}}
        \STATE{Compute $\mathbf{dM}_{t} = (\mathbf Q_t)^{\top}  \mathbf {dO}_t$.}
        \STATE{\color{magenta}{Communicate $[\mathbf{dM}]_{1}^T = \texttt{AllGather}([\mathbf{dM}]_{1}^T)$.}}
        \STATE{\color{cyan}{Compute 
        $\mathbf {dQ}_{\mathrm{t, intra}}=[(\mathbf {dO}_t \mathbf V_t^{\top}) \odot \mathbf \Psi] \mathbf{K}_t$.}}
        \STATE{\color{cyan}{Compute
        $\mathbf{dK_{\mathrm{t, intra}}}=[(\mathbf {dO}_t \mathbf V_t^{\top}) \odot \mathbf \Psi ]^{\top} \mathbf{Q}_t$.}}
        \STATE{\color{cyan}{Compute
        $\mathbf{dV_{\mathrm{t, intra}}}=[(\mathbf Q_t \mathbf K_t^{\top}) \odot \mathbf \Psi ]^{\top} \mathbf{dO}_t$.}}
        \STATE{\color{cyan}{Compute
        $\mathbf{dQ}_{\mathrm{t, inter}}=\mathbf{dO}_t \mathbf{M}_{1:t-1}^{\top}$.}}
        \STATE{Compute suffix sum $\mathbf{dM}_{t+1:T} = \texttt{SuffixSum}([\mathbf{dM}]_{t+1}^T)$.}
        \STATE{Compute
        $\mathbf{dK_{\mathrm{t, inter}}}={\mathbf V_t}\mathbf{dM}_{t+1:T}^{\top}$.}
        \STATE{Compute
        $\mathbf{dV_{\mathrm{t, inter}}}=\mathbf K_t \mathbf{dM}_{t+1:T}$.}
        \STATE{Combine intra- and inter-chunk parts of $\mathbf {dQ}_t, \mathbf {dK}_t, \mathbf {dV}_t$
        \begin{equation*}
        \begin{aligned}
        \mathbf {dQ}_t=\mathbf{dQ}_{t,\mathrm{intra}} + \mathbf{dQ}_{t,\mathrm{inter}}, \quad
        \mathbf {dK}_t=\mathbf {dK}_{t,\mathrm{intra}} +\mathbf {dK}_{t,\mathrm{inter}}, \quad
        \mathbf {dV}_t=\mathbf {dV}_{t,\mathrm{intra}} +\mathbf {dV}_{t,\mathrm{inter}}.
        \end{aligned}
        \end{equation*}}
    \ENDFOR
    \STATE{return $\mathbf {dQ}=[\mathbf {dQ}_t]_1^T$, $\mathbf{dK}=[\mathbf{dK}_t]_1^T$, $\mathbf{dV}=[\mathbf{dV}_t]_1^T$.}
\end{algorithmic}
\end{algorithm*}

\subsection{LASP-1 Algorithms}
\label{app: lasp-1 algo}

See Algorithm~\ref{algo: lasp1 forward} and Algorithm~\ref{algo: lasp1 masking forward}.

\begin{algorithm}[ht]
\caption{LASP-1 w/o Masking}
\label{algo: lasp1 forward}
\begin{algorithmic}[1]
\STATE{\textbf{Input:} input sequence $\mathbf X$, distributed world size $W$, sequence parallel size $T=W$.}
\STATE {Distribute input $\mathbf{X} = [\mathbf{X}_t]_{1}^{T}$.}
\FOR{chunk $t \in \{1, \cdots, T\}$ at rank $i \in \{1, \cdots, W\}$ \textbf{in parallel}}
    \STATE Compute $\mathbf{Q}_t = \mathbf{X}_t \mathbf{W}_Q$, 
    $\mathbf{K}_t = \mathbf{X}_t \mathbf{W}_K$, 
    $\mathbf{V}_t = \mathbf{X}_t \mathbf{W}_V$.
    \STATE Compute $\mathbf{M}_t = \mathbf{K}_t^\top \mathbf{V}_t$.
\ENDFOR
\FOR{chunk $t \in \{1, \cdots, T\}$ at rank $i \in \{1, \cdots, W\}$ \textbf{sequentially}}
    \STATE \texttt{Recv} activation $\mathbf{M}_{t-1}$ from rank $(i-1)$. Save $\mathbf{M}_{t-1}$ in memory for backward computation.
    \STATE Compute $\mathbf{O}_{t} = \mathbf{Q}_t \mathbf{M}_{t-1}$.
    \STATE Update $\mathbf{M}_t = \mathbf{M}_{t-1} + \mathbf{K}_t^\top \mathbf{V}_t$.
    \STATE \texttt{Send} activation $\mathbf{M}_t$ to rank $(i+1)$.
\ENDFOR
\STATE \textbf{return} $\mathbf{O} = [\mathbf{O}_t]$ with ${t \in \{1, \cdots, T\}}$.
\end{algorithmic}
\end{algorithm}

\begin{algorithm}[ht]
\caption{LASP-1 w/ Masking}
\label{algo: lasp1 masking forward}
\begin{algorithmic}[1]
\STATE{\textbf{Input:} input sequence $\mathbf X$, distributed world size $W$, sequence parallel size $T=W$.}
\STATE {Distribute input $\mathbf{X} = [\mathbf{X}_t]_{1}^{T}$.}
\STATE Initialize mask matrix $\mathbf{\Psi}$, where 
$\mathbf{\Psi}_{ij} = 1$ if $i \geq j$, and $\mathbf{\Psi}_{ij} = -\infty$ if $i < j$.
\FOR{chunk $t \in \{1, \cdots, T\}$ at rank $i \in \{1, \cdots, W\}$ \textbf{in parallel}}
    \STATE Compute $\mathbf{Q}_t = \mathbf{X}_t \mathbf{W}_Q$, 
    $\mathbf{K}_t = \mathbf{X}_t \mathbf{W}_K$, 
    $\mathbf{V}_t = \mathbf{X}_t \mathbf{W}_V$.
    \STATE Compute $\mathbf{M}_t = (\mathbf{K}_t)^\top \mathbf{V}_t$.
    \STATE Compute $\mathbf{O}_{\mathrm{t, intra}} = [(\mathbf{Q}_t \mathbf{K}_t^\top) \odot \mathbf{\Psi}] \mathbf{V}_t$.
\ENDFOR
\FOR{chunk $t \in \{1, \cdots, T\}$ at rank $i \in \{1, \cdots, W\}$ \textbf{sequentially}}
    \STATE \texttt{Recv} activation $\mathbf{M}_{t-1}$ from rank $(i-1)$. Save $\mathbf{M}_{t-1}$ in memory for backward computation.
    \STATE Compute $\mathbf{O}_{\mathrm{t, inter}} = \mathbf{Q}_t \mathbf{M}_{t-1}$.
    \STATE Compute $\mathbf{O}_t = \mathbf{O}_{\mathrm{t, intra}} + \mathbf{O}_{\mathrm{t, inter}}$.
    \STATE Update $\mathbf{M}_t = \mathbf{M}_{t-1} + \mathbf{K}_t^\top \mathbf{V}_t$.
    \STATE \texttt{Send} activation $\mathbf{M}_t$ to rank $(i+1)$.
\ENDFOR
\STATE \textbf{return} $\mathbf{O} = [\mathbf{O}_t]$ with ${t \in \{1, \cdots, T\}}$.
\end{algorithmic}
\end{algorithm}

\subsection{AllGather-based Context Parallelism}
\label{app: AllGather-based Context Parallelism}

See Algorithm~\ref{algo:allgather_context_parallelism}.

\begin{algorithm}[ht]
\caption{AllGather-based Context Parallelism}
\label{algo:allgather_context_parallelism}
\begin{algorithmic}[1]
\STATE{\textbf{Input:} input sequence $\mathbf X$, distributed world size $W$, sequence parallel size $T=W$.}
\STATE Distribute $\mathbf{X} = [\mathbf{X}_t]_1^T$.
\FOR{chunk $t \in \{1, \cdots, T\}$ on ranks $\{1, \cdots, W\}$ in parallel}
    \STATE Calculate $\mathbf{Q}_t = \mathbf{X}_t \mathbf{W}_Q$, $\mathbf{K}_t = \mathbf{X}_t \mathbf{W}_K$, $\mathbf{V}_t = \mathbf{X}_t \mathbf{W}_V$.
    \STATE Communicate $[\mathbf{K}_t]_1^T = \texttt{AllGather}([\mathbf{K}_t]_1^T)$ and $[\mathbf{V}_t]_1^T = \texttt{AllGather}([\mathbf{V}_t]_1^T)$.
    \STATE Concatenate $\mathbf{K} = \texttt{Concat}([\mathbf{K}_t]_1^T)$ and $\mathbf{V} = \texttt{Concat}([\mathbf{V}_t]_1^T)$.
    \STATE Compute $\mathbf{O}_{t} = \texttt{Softmax}(\mathbf{Q}_t \mathbf{K}^\top / \sqrt{d})\mathbf{V}$.
\ENDFOR
\STATE \textbf{return} $\mathbf{O} = [\mathbf{O}_t]_1^T$.
\end{algorithmic}
\end{algorithm}



\subsection{Compatibility}

\subsubsection{Hybrid Parallelism}
\label{subsec: hybrid parallel}



LASP-2 enables the selection of a sequence parallel size that is smaller and divisible by the distributed world size. This setup splits the input data along both the batch and sequence dimensions, a parallelization strategy known as data-sequence hybrid parallelism. The ZeRO-series optimizers~\citep{rajbhandari2020zero} and FSDP~\citep{zhao2023pytorch} offer methods for distributing model states such as optimizer states, gradients, and model parameters across all GPUs in the distributed system. As these techniques are variants of data parallelism, they integrate seamlessly with LASP. Their primary objective of minimizing the memory footprint of model states complements LASP-2’s specific focus on reducing activation memory on each GPU, making the training of large-scale models that handle long sequence lengths significantly more manageable.

LASP-2 also offers support for both tensor parallelism (TP) and pipeline parallelism (PP). In the case of TP, its integration with LASP-2 is straightforward and efficient. Linear attention layers apply TP to break down matrix operations across both intra-chunk and inter-chunk computations. At the same time, the MLP layers are processed as usual under TP, without any modification. When LASP-2 is paired with PP, instead of using traditional micro-batches, it substitutes them with sub-sequences extracted from the mini-batch. One key difference from standard PP is that each device locally and specifically stores the intermediate states, $\mathbf{M}_t$ during the forward pass and $\mathbf{dM}_t$ during the backward pass without communicating these states to other devices.


\subsubsection{Variable Length}
During pretraining, the batch typically contains sequences of uniform length. However, when finetuning or during inference, the model might encounter input sequences of varying lengths. A straightforward solution to address this is to right-pad all sequences in a batch to match the length of the longest sequence. Unfortunately, this method can be inefficient, especially when the lengths differ significantly across sequences. For standard transformers, more sophisticated approaches have been developed to handle this challenge. These include techniques like load-balancing across GPUs without padding~\citep{zeng2022boosting, zhai2023bytetransformer} or packing multiple sequences into a single batch and adjusting the attention mask accordingly~\citep{ding2024fewer, pouransari2024dataset}. LASP-2 can manage variable sequence lengths efficiently by treating the entire batch as a single long sequence, streamlining the process without requiring padding.

\subsection{Additional Experiment Results}
\label{app: additional exp}

\subsubsection{Bidirectional Language Modeling Task}

To evaluate on the bidirectional language modeling task, we take RoBERTa as the base model and replace its standard attention modules with Basic Linear Attention, train it on 4 A100 GPUs for 50K iterations with a total input sequence length of 2048. As the results shown in Table~\ref{tab: bidirectional loss_comparison}, LASP-2 with Basic Linear Attention is able to reach an approximate convergence performance with Ring Attention on the standard attention based model.

\begin{table}[ht]
\centering
\small
\caption{\textbf{Convergence Performance on Bidirectional Language Modeling Task.} Both training and validation loss values are reported.}
\label{tab: bidirectional loss_comparison}
\begin{tabular}{lcc}
\toprule
\textbf{Model} & \textbf{Training Loss} & \textbf{Validation Loss} \\
\midrule
RoBERTa Baseline (Ring Attention) & 1.815 & 1.957 \\
\midrule
RoBERTa with Basic Linear Attention (LASP-2) & 1.813 & 1.957 \\
\bottomrule
\end{tabular}
\end{table}




\subsubsection{Ablation Study on Hybrid Ratio}
We provide ablation results on the hybrid ratio of hybrid models. Let "L" denotes linear Transformer layers and "N" denotes normal Transformer layers. The hybrid models evaluated here have architectures of: 0 Hybrid: "LLLL LLLL LLLL LLLL"; 1/8 Hybrid: "LLLL LLLN LLLL LLLN"; 1/4 Hybrid: "LLLN LLLN LLLN LLLN"; 1/2 Hybrid: "LNLN LNLN LNLN LNLN".
Comparing with the Llama3-1B baseline using standard attention, whose loss value is 2.759, it shows that higher hybrid ratios tend to lead better convergence performance, but sometimes, a moderate hybrid ratio may reach a better result.

\begin{table}[ht]
\centering
\small
\caption{\textbf{Ablation Study on Hybrid Ratio in Hybrid Models.} Loss values are reported in the Table. Note that pure linear models use LASP-2, while hybrid models use LASP-2H.}
\label{tab: hybrid ratio}
\begin{tabular}{lcccc}
\toprule
\textbf{Linear Sequence Modeling Module} & \textbf{0 Hybrid (Pure Linear Model)} & \textbf{1/8 Hybrid} & \textbf{1/4 Hybrid} & \textbf{1/2 Hybrid} \\
\midrule
Basic Linear Attention & 2.892 & 2.826 & 2.824 & 2.775 \\
\midrule
Lightning Attention & 2.848 & 2.756 & 2.750 & 2.742 \\
\midrule
Retention & 2.855 & 2.757 & 2.758 & 2.748 \\
\midrule
GLA & 2.845 & 2.751 & 2.754 & 2.753 \\
\bottomrule
\end{tabular}
\end{table}

\subsubsection{Ablation Study on Varying Sizes of Gathering}
We have conducted ablation study on varying sizes of gathering memory states. Considering a batch size of 1, in the Linear-Llama3-1B model (with 16 heads and hidden dimension of 2048), the tensor shape of each memory state is $[1, 16, 2048, 2048]$. We use 64 GPUs and a sequence lenghth of 1024K, repeat each test 10 times and report their mean values. We change the split size of gathering memory states and present the LASP-2 throughput results in Table~\ref{tab: throughput gathering size}. It can be seen that smaller split size (i.e., more number of splits) tends to lead lightly slower throughput. The results show that the utilization of all-gather operation is not the only reason of efficiency enhancement. The communication manner as well as the computational workflow reorganization plays an important role.

\begin{table}[ht]
    \centering
    \small
    \caption{\textbf{Throughput Results (tokens/sec) on Varying Split Sizes of Gathering.} Linear-Llama3-1B model (with 16 heads and hidden dimension of 2048) is used.}
    \label{tab: throughput gathering size}
    \begin{tabular}{lccccc}
        \toprule
        \textbf{Split Size of Gathering} & 2048 & 512 & 128 & 32 \\
        \midrule
        \textbf{Number of Splits} & 1 & 4 & 16 & 64 \\
        \midrule
        \textbf{Throughput} & 486183 & 486166 & 486169 & 486158 \\
        \bottomrule
    \end{tabular}
\end{table}

\subsubsection{Quantitative Scalability Results}
See Table~\ref{tab:quantitative} in next page.

\begin{table*}[htbp]
    \centering
    \small    
    \caption{\textbf{Quantitative Scalability Results of LASP-2 on Throughput (tokens/sec) and Memory Usage Per GPU (GB).} Experiments are performed on Linear-Llama3-1B, scaling sequence length from 2K to 4096K.}
        \begin{tabular}{cccc}
        \toprule
            \textbf{Sequence Length} & \textbf{Number of GPUs} &\textbf{Throughput}      &\textbf{Memory Usage Per GPU}    \\ \midrule
            \multirow{4}{*}{\textbf{2K}}
                        & 16  & 1254 & 25.6 \\
                        & 32  & 1209 & 25.6 \\
                        & 64  & 1285 & 25.6 \\
                        & 128 & 1205 & 25.6 \\  \midrule
            \multirow{4}{*}{\textbf{4K}}
                        &16  & 2478 & 25.6 \\
                        &32  & 2446 & 25.6 \\
                        &64  & 2327 & 25.6 \\
                        &128 & 2344 & 25.6 \\  \midrule
            \multirow{4}{*}{\textbf{8K}} 
                        & 16   & 4835 & 25.6 \\
                        & 32   & 4784 & 25.6 \\ 
                        & 64   & 4693 & 25.6 \\
                        & 128  & 4678 & 25.6 \\ \midrule
            \multirow{4}{*}{\textbf{16K}} 
                        & 16   & 9530 & 25.6 \\
                        & 32   & 9494 & 25.6 \\
                        & 64   & 9305 & 25.6 \\
                        & 128  & 9313 & 25.6 \\ \midrule
            \multirow{4}{*}{\textbf{32K}}
                        & 16   & 18105 & 28.7 \\
                        & 32   & 17755 & 25.6 \\
                        & 64   & 17835 & 25.6 \\
                        & 128  & 17807 & 25.6 \\ \midrule
            \multirow{4}{*}{\textbf{64K}}
                        & 16   & 35507 & 33.8 \\
                        & 32   & 34240 & 28.7 \\
                        & 64   & 34118 & 25.6 \\
                        & 128  & 33344 & 25.6 \\ \midrule
            \multirow{4}{*}{\textbf{128K}} 
                        & 16   & 68406 & 40.2 \\
                        & 32   & 68545 & 33.8 \\
                        & 64   & 67344 & 28.7 \\
                        & 128  & 66811 & 25.6 \\ \midrule
            \multirow{4}{*}{\textbf{256K}}
                        & 16   & 135635 & 57.8 \\
                        & 32   & 132605 & 40.2 \\
                        & 64   & 130215 & 33.8 \\ 
                        & 128  & 131550 & 28.7 \\ \midrule
            \multirow{4}{*}{\textbf{512K}}
                        & 16   & OOM &OOM     \\
                        & 32   & 250586 & 57.8  \\
                        & 64   & 245353 & 40.2   \\
                        & 128  & 233442 & 33.8   \\ \midrule
            \multirow{4}{*}{\textbf{1024K}}
                        & 16   & OOM &OOM      \\
                        & 32   & OOM &OOM      \\
                        & 64   & 442221 & 57.8   \\
                        & 128  & 416465 & 40.2   \\ \midrule
            \multirow{4}{*}{\textbf{2048K}}
                        & 16   & OOM & OOM       \\
                        & 32   & OOM & OOM         \\
                        & 64   & OOM & OOM   \\
                        & 128  & 769030 & 57.8   \\ \midrule
            \multirow{4}{*}{\textbf{4096K}}
                        & 16   & OOM & OOM     \\
                        & 32   & OOM & OOM        \\
                        & 64   & OOM & OOM        \\
                        & 128  & OOM & OOM     \\
        \bottomrule             
        \end{tabular}
 \label{tab:quantitative}
\end{table*}

\end{document}